# Face Verification via learning the kernel matrix

Ning Yuan, Xiao-Jun Wu, He-Feng Yin

School of IoT Engineering, Jiangnan University, Wuxi, China

**Abstract**   The kernel function is introduced to solve the nonlinear pattern recognition problem. The advantage of a kernel method often depends critically on a proper choice of the kernel function. A promising approach is to learn the kernel from data automatically. Over the past few years, some methods which have been proposed to learn the kernel have some limitations: learning the parameters of some prespecified kernel function and so on. In this paper, the nonlinear face verification via learning the kernel matrix is proposed. A new criterion is used in the new algorithm to avoid inverting the possibly singular within-class which is a computational problem. The experimental results obtained on the facial database XM2VTS using the Lausanne protocol show that the verification performance of the new method is superior to that of the primary method Client Specific Kernel Discriminant Analysis (CSKDA). The method CSKDA needs to choose a proper kernel function through many experiments, while the new method could learn the kernel from data automatically which could save a lot of time and have the robust performance.

## 1. Introduction

Face Verification is the problem which judges the face of the image is whether or not the assigner[1].

Feature extraction is a key step for face verification[2-7]. Turk et al proposed the classical eigen face method[8], which extract feature using the principal component analysis (PCA) and get the better result. However this method just considers the second order statistical information and mislays the data's high order information and the nonlinear relation [9]. Through the research, we find that the high order statistical information include the pixels nonlinear relation of the image edge or the curve [10]. So the kernel operation is introduced to solve the problem, and then Scholkopf[10] proposed the nonlinear spread kernel PCA and Mika[11] proposed the nonlinear spread of LDA

In the application of face verification, Kittler introduced the Client Specific to the face verification based on the LDA in order to get the better performance [12-15]. After that, Yuan, Wu and Kittler introduced the kernel operation to extend the LDA method based on the Client Specific and proposed some methods about the Client Specific [16-17] to get the better results.

However the advantage of a kernel method often depends critically on a proper choice of the kernel function. Early work on kernel learning is limited to learning the parameters of some prespecified kernel function [18]。More recent work has gone beyond kernel parameter learning by learning the kernel itself in a more nonparametric manner. In practice, since we work with data sets of finite size, we can learn the kernel matrix corresponding to a given date set instead of learning the kernel function [19]-[24].

The method kernel matrix learning proposed by Dit-Yan et al [24] is based on optimizing the Fisher criterion and use it to classification get the better result. The paper proposed face verification via learning the kernel matrix based on that and be applied to CSKDA[16]. The experimental results obtained on the facial database show that the verification performance of the new method is superior to that of the primary method Client Specific Kernel Discriminant Analysis (CSKDA). The method CSKDA need to choose a proper kernel function through many experiments, while the new method could learn the kernel from data automatically which could save a lot of time and have the robust performance. In this paper the second part introduces the method CSKDA; The third part writes up the paper's method; The forth part gives the experimental results and analysis, and finally we get the conclusion.

## 2. CSKDA method

A set of $n$ training face images $x_i, i=1,\cdots n$ is available in the face database. Each image is defined as a vector of length $M$, i.e. $x_i \in R^M$, where $M$ is the face image size and $R^M$ denotes a M-dimensional real space. It is assumed that each image belongs to one of the C classes denoted by $\omega_1, \omega_2, \cdots \omega_c$. The i-th calss include $n_i$ samples. A nonlinear mapping is associated with the kernel discriminant analysis. Let $\phi: x \in R^M \to \phi(x) \in F$ be a nonlinear mapping from the input space to a high-dimensional feature space F, where different classes of faces are linear separable with great potentiality. Let us now consider the problem of discriminating class $\omega_i$ from all the other classes. In the context of the face verification problem this corresponds to discriminating between i-th client and imposters modeled by all the other clients in the training data set. Given the mean vector of i-th class as

$$m_i = \frac{1}{n_i} \sum_{j=1}^{n_i} \phi(x_j) \qquad (1)$$

The Fisher discriminant function can be defined as

$$J(v) = \frac{v^T S_{bi} v}{v^T S_{wi} v} \qquad (2)$$

$S_{bi}$ is the between-class scatter matrix and $S_{wi}$ is the within-class scatter matrix. The solution to the problem can be found easily as $v = S_{wi}^{-1} \mu_i$, that is the $i$-th Fisher subspace[16]. Then using the testing samples projected to the client specific Fisher space to classify.

## 3. Learning the kernel matrix

A set of $n$ training face images $x_i$, $E$ evaluation images and $I$ testing images is available in the face database. In the samples some is in the $c$ classes and some is not in the classes, so there are samples $N=n+E+I$. A nonlinear mapping $\phi$ is associated with the kernel discriminant analysis.

3.1 Spectral variants of kernel matrix[24]

Let $K=[k(x_i, x_j)]_{N \times N}$ denote the kernel matrix formed by the $N$ points for some chosen kernel function $k(\cdot, \cdot)$. We express the spectral decomposition of $K$ as

$$K = \sum_{r=1}^{p} \lambda_r v_r v_r^{\mathrm{T}} = \sum_{r=1}^{p} \lambda_r K_r \qquad (3)$$

Where $\lambda_r$ $(r = 1, \cdots, p)$ are the $p$ positive eigenvalues of $K$ sorted in a monotonically decreasing order and $v_r$ $(r = 1, \cdots, p)$ are the corresponding normalized eigenvectors. $K_r$ are base kernel matrices of rank-one. We define a parameterized family of kernel matrices as

$$K_\mu = \sum_{r=1}^{p} \mu_r^2 v_r v_r^{\mathrm{T}} = \sum_{r=1}^{p} \mu_r^2 K_r \qquad (4)$$

Where $\mu = (\mu_1, \cdots, \mu_p)^{\mathrm{T}}$ denotes $p$ coefficients for specifying the spectral variants.

### 3.2 Optimization[19]

The Fisher discriminant function can be defined as
$$J(v) = \frac{v^{\mathrm{T}} S_b v}{v^{\mathrm{T}} S_w v} \qquad (5)$$

Where $S_b$ is the between-class scatter matrix and $S_w$ is the within-class scatter matrix

$$S_b = \frac{1}{n} \sum_{i=1}^{c} n_i (m_i - m)(m_i - m)^{\mathrm{T}} \qquad (6)$$

$$S_w = \frac{1}{n} \sum_{i=1}^{c} \sum_{j=1}^{n_i} (\phi(x_j) - m_i)(\phi(x_j) - m_i)^{\mathrm{T}} \qquad (7)$$

$m_i$ is the mean vector of i-th class, $m$ is the mean vector of all training images

$$m = \frac{1}{n} \sum_{i=1}^{n} \phi(x_i) \qquad (8)$$

However, we can see from equation (5) that the inverse of the within-class scatter matrix should be calculated for each client. Here we use the trace of a scatter matrix to quantify its scatter. Let the (i,j)-th entry of $K_\mu$ be expressed as $(K_\mu)_{ij} = b_i^{\mathrm{T}} K_\mu b_j$, where $b_i$ is the i-th column of the $n \times n$ identity matrix.

We use $Tr(S_b)$ and $Tr(S_w)$ to characterize the between-class and within-class scatter matrix respectively:

$$Tr(S_b) = \frac{1}{n}\sum_{i=1}^{c} n_i(m_i - m)^T(m_i - m)$$

$$= \frac{1}{n}\sum_{i=1}^{c} n_i m_i^T m_i - m^T m$$

$$= \frac{1}{n}\left[\sum_{i=1}^{c}\sum_{k=1}^{n_i}\frac{1}{n_i}(K_\mu)_{jk} - \sum_{j,k=1}^{n}\frac{1}{n}(K_\mu)_{jk}\right]$$

$$= \frac{1}{n}\sum_{j,k=1}^{n}(\sum_{i=1}^{c}a_{jk}^i - \frac{1}{n})b_j^T K_\mu b_k$$

$$= \sum_{r=1}^{p}\mu_r^2\left[\frac{1}{n}\sum_{j,k=1}^{n}(\sum_{i=1}^{c}a_{jk}^i - \frac{1}{n})b_j^T K_r b_k\right]$$

$$= \sum_{r=1}^{p}\mu_r^2 f_r$$

$$= \mu^T D_b \mu \qquad (9)$$

where

$$a_{jk}^i = \begin{cases} \frac{1}{n_i} & x_j, x_k \in w_i \\ 0 & x_j, x_k \notin w_i \end{cases} \qquad (10)$$

$$f_r = \frac{1}{n}\sum_{j,k=1}^{n}(\sum_{i=1}^{c}a_{jk}^i - \frac{1}{n})b_j^T K_r b_k$$

$$= \frac{1}{n}\sum_{j,k=1}^{n}(\sum_{i=1}^{c}a_{jk}^i - \frac{1}{n})b_j^T v_r v_r^T b_k \qquad (11)$$

$$D_b = diag(f_1, \cdots f_p) \qquad (12)$$

Similarly, we can rewrite that

$$Tr(S_w) = \frac{1}{n}\sum_{i=1}^{c}\sum_{j=1}^{n_i}(\phi(x_j) - m_i)^T(\phi(x_j) - m_i)$$

$$= \frac{1}{n}\sum_{i=1}^{c}\left(\sum_{j=1}^{n_i}\phi(x_j)^T\phi(x_j) - n_i m_i^T m_i\right)$$

$$= \frac{1}{n}\left[\sum_{j=1}^{n}(K_\mu)_{jj} - \sum_{i=1}^{c}\sum_{k=1}^{n_i}\frac{1}{n_i}(K_\mu)_{jk}\right]$$

$$= \frac{1}{n}\sum_{j,k=1}^{n}(\delta_{jk} - \sum_{i=1}^{c}a_{jk}^i)b_j^T K_\mu b_k$$

$$= \sum_{r=1}^{p}\mu_r^2\left[\frac{1}{n}\sum_{j,k=1}^{n}(\delta_{jk} - \sum_{i=1}^{c}a_{jk}^i)b_j^T K_r b_k\right]$$

$$= \sum_{r=1}^{p}\mu_r^2 g_r$$

$$= \mu^T D_w \mu$$

(13)

where

$$\delta_{jk} = \begin{cases} 1 & j = k \\ 0 & j \neq k \end{cases} \quad (14)$$

$$g_r = \frac{1}{n}\sum_{j,k=1}^{n}(\delta_{jk} - \sum_{i=1}^{c}a_{jk}^i)b_j^T K_r b_k$$
$$= \frac{1}{n}\sum_{j,k=1}^{n}(\delta_{jk} - \sum_{i=1}^{c}a_{jk}^i)b_j^T v_r v_r^T b_k \quad (15)$$

$$D_w = diag(g_1,\cdots,g_p) \quad (16)$$

One possible optimality criterion for maximization is $Tr(S_b)/Tr(S_w)$ and the best solution is the eigenvector corresponding to the largest eigenvalue of $D_w^{-1}D_b$. This implies that the spectral variant solution degenerates to having only one base kernel. Apparently this is not what we want[24]. An alternative approach is to regard the maximization of $Tr(S_b)/Tr(S_w)$ as a nonlinear fractional programming problem [25], we define the following criterion function:

$$Q(\mu) = Tr(S_b) - \alpha Tr(S_w) = \mu^T(D_b - \alpha D_w)\mu \quad (17)$$

Where $\alpha > 0$ is a parameter that can be determined. The optimal value of (17) is given by

$$\mu = \frac{\beta(D_b - \alpha)^{-1}\theta}{\theta^T(D_b - \alpha)^{-1}\theta} \quad (18)$$

$\theta$ is a column vector of ones and $\beta = \sum_{r=1}^{p}\sqrt{\lambda_r}$。

The learned kernel matrix $K_\mu$ can then be used to face verification of CSKDA.

3.3 application of kernel matrix in CSKDA
3.3.1 dimensionality reduction

In small sample cases where the dimensionality of the data exceeds the cardinality of the training set, LDA has to be preceded by a dimensionality reduction in order to avoid the problem of rank deficiency of the population scatter matrix. The between-class scatter matrix $S_b$ is defined as

$$S_b = \frac{1}{n}\sum_{i=1}^{c}n_i(m_i - m)(m_i - m)^T$$
$$= \sum_{i=1}^{c}\left(\sqrt{\frac{n_i}{n}}(m_i - m)\right)\left(\sqrt{\frac{n_i}{n}}(m_i - m)\right)^T \quad (19)$$
$$= P_b P_b^T$$

Let $K_t = K_\mu(1:n,1:n)$, and the between-class scatter matrix in the kernel space can be rewritten as follows

$$P_b^T P_b = \frac{1}{n} B(A_{nc}^T \cdot K_t \cdot A_{nc} - \frac{1}{n}(A_{nc}^T \cdot K_t \cdot I_{nc}) \quad (20)$$
$$- \frac{1}{n}(I_{nc}^T \cdot K_t \cdot A_{nc}) + \frac{1}{n^2}(I_{nc}^T \cdot K_t \cdot I_{nc})) \cdot B$$

Where $B = diag[\sqrt{n_1} \cdots \sqrt{n_c}]$, $I_{nc}$ is a $n \times c$ matrix of ones, $A_{nc} = diag[a_{n_1} \cdots a_{n_C}]$ is a $n \times c$ block diagonal matrix, and $a_{n_i}$ is a $n_i \times 1$ vector with all terms equal to $\frac{1}{n_i}$.

Let $\lambda_i$ and $e_i$ be the i-th eigenvalue and corresponding eigenvector of $P_b^T P_b$, sorted in decreasing order of eigenvalues. We only use its first $m\_b$ eigenvectors:

$$m\_b = rank(S_b) \le \min(n, c-1)$$

Then the eigenvectors of $S_b$ is

$$T = [t_1, t_2, \cdots t_{m\_b}] = P_b E_m \quad E_m = [e_1, e_2, \cdots, e_{m\_b}]$$

Let $U_b = diag(\lambda_1, \cdots, \lambda_{m\_b})$ and further let $U = TU_b^{-\frac{1}{2}}$

### 3.3.2 calculating projecting vectors

Projecting all the training samples into the subspace spanned by U, we have $y_i$. Let us now consider the problem of discriminating class $\omega_i$ from all the other classes. In the context of the face verification problem this corresponds to discriminating between i-th client and imposters modeled by all the other clients in the training data set. In this two class scenario, LDA involves finding one dimensional feature space. Given the mean vector of i-th class as:

$$\mu_i = \frac{1}{n_i} \sum_{j=1}^{n_i} y_j \quad (21)$$
$$= (E_m U_b^{-\frac{1}{2}})^T \frac{1}{n_i} \sum_{j=1}^{n_i} \left( P_b^T \phi(y_j) - \frac{1}{n} \sum_{k=1}^{n} P_b^T \phi(y_k) \right)$$

and assuming the population mean to be zero, it can be shown that the mean vector of impostors of *i-th* class is

$$\mu_\Omega = -\frac{n_i}{n - n_i} \mu_i \quad (22)$$

In the rest of the section, we propose to utilize an equivalent Fisher criterion function [11]

$$J(v) = \frac{v^T S_{bi} v}{v^T S_t v} \quad (23)$$

Where $S_{bi}$ is the between-class scatter matrix of the client and tne imposter models, $S_t$ is the population scatter matrix.

$$S_{bi} = \frac{n_i}{n - n_i} \mu_i \mu_i^T \quad (24)$$

$$S_t = \frac{1}{n} \sum_{i=1}^{n} \phi(y_i) \phi(y_i)^T \quad (25)$$

The solution to the problem can be found easily as

$$v_i = S_t^{-1} \mu_i \quad (26)$$

Thus the overall client i specific discriminant transformation $a_i$, which defines the client specific fisher face of the claimed identity, is given as

$$a_i = v_i \quad (27)$$

3.3.3 Classification

The testing image $z$, project to the subspace $\tilde{z} = (E_m U_b^{-\frac{1}{2}} a_i)^T (P_b^T \phi(z))$, where

$$P_b^T \phi(z) = \frac{1}{\sqrt{n}} \cdot B \cdot \left( A_{nc}^T \cdot r(\phi(z)) - \frac{1}{n} I_{nc}^T \cdot r(\phi(z)) \right) \quad r(\phi(z)) = [\phi_{11}^T \phi(z), \phi_{12}^T \phi(z), \cdots \phi_{cn_i}^T \phi(z)]^T$$
(28)

While in the learning kernel matrix $K_\mu$, we can get the $r(\phi(z))$.

The mean vector of i-th class $m_i$ project to the subspace $\tilde{m}_i$, the imposter mean vector of i-th class $m_\Omega$ project to the subspace $\tilde{m}_\Omega$.

The classification based on client model: the distance between the testing sample and the i-th mean vector is defined as: $d_c = |\tilde{z} - \tilde{u}_i|$. If the distance exceeds a predefined threshold $t_c$ the claim is rejected, otherwise the claimed identity is accepted.

The classification based on imposter model: the distance between the testing sample and the i-th imposter mean vector is defined as: $d_i = |\tilde{z} - \tilde{m}_\Omega|$. If the distance exceeds a predefined threshold $t_c$ the claim is accepted, otherwise the claimed identity is rejected.

## 4. Experimental results and analysis

In order to test the performance of the proposed algorithm, face verification experiments have been conducted on the XM2VTS database, which is a multi-modal database consisting of video sequences of talking faces recorded for 295 subjects at one month intervals. The data has been recorded in 4 sessions with 2 shots taken per session [14]. From each session two facial images have been extracted to create an experimental face database of size 55 × 51. Figure 1 shows examples of images in XM2VTS.

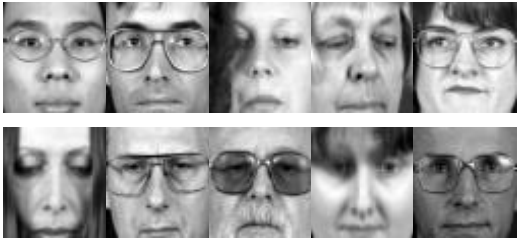

Fig 1 Part of the extracted images in

The experimental protocol (known as Lausanne evaluation protocol) divides the data set into 200 clients and 95 impostors[14]. Within the protocol, the verification performance is measured using false acceptance and false rejection rates. The operating point where these two error rates equal each other is typically referred to as the equal error rate (EER) point. All the results were obtained using histogram equalization (HEQ) in conjunction with a global threshold determined by the EER point.

This paper use the minimize distance classifier. In the evaluation process, we modified the threshold to get the FAR and FRR same in order to require the final threshold. And then the paper use the final threshold to testing the method.

The comparison between the new method and CSKDA use the Client model and the imposter model classification and make them to OnC and OnI for short. In the table, TER=FAR+FRR, denotes the total error rate.

Because of the kernel operation, we should consider choosing the kernel function and its parameters. Tab.1 use the Polynomial kernel function $K(x,y) = (a(x \cdot y) + b)^d$, the results as follows with the different parameters:

Tab.1 Performance comparison of CSKDA with the new method using Polynomial kernel function On XM2VTS

| Method | parameters | | | Client Imposter | evaluation | | testing | | |
|---|---|---|---|---|---|---|---|---|---|
| | a | b | d | | FAR | FRR | FAR | FRR | TER |
| CSKDA | 0.0001 | 1 | 2 | OnC | 4.25 | 4.25 | 4.12 | 6 | 10.12 |
| | | | | OnI | 1.55 | 1.5 | 1.1 | 3 | 4.1 |
| The new method | 0.0001 | 1 | 2 | OnC | 5.5 | 5.5 | 5.21 | 7 | 12.21 |
| | | | | OnI | 1.75 | 1.75 | 1.29 | 3.25 | 4.54 |
| | 0.0001 | 0 | 2 | OnC | 4.25 | 4.25 | 4.10 | 5.5 | 9.60 |
| | | | | OnI | 1.25 | 1.25 | 1.03 | 3 | 4.03 |
| | 10 | 1 | 2 | OnC | 4.25 | 4.25 | 4.10 | 5.5 | 9.60 |
| | | | | OnI | 1.25 | 1.25 | 1.03 | 3 | 4.03 |
| | 5 | 2 | 4 | OnC | 3.88 | 4 | 3.95 | 6 | 9.95 |
| | | | | OnI | 1.5 | 1.5 | 1.29 | 2.75 | 4.04 |

Tab.2 use the RBF kernel function $K(x,y) = e^{-\frac{\|x-y\|}{\sigma^2}}$, the results as follows with the different parameters:

Tab.2 Performance comparison of CSKDA with the new method using RBF kernel function On XM2VTS

| Method | parameters | Client | evaluation | testing |
|---|---|---|---|---|

|  |  | Imposter | FAR | FRR |  | FAR | FRR |
|---|---|---|---|---|---|---|---|
| CSKDA | 20 | OnC | 5.42 | 5.5 | 5.30 | 7.25 | 12.55 |
|  |  | OnI | 2.02 | 2 | 2.16 | 1.5 | 3.66 |
| The new method | 5 | OnC | 5.75 | 5.75 | 5.66 | 7.25 | 12.91 |
|  |  | OnI | 1.75 | 1.75 | 2.07 | 1.5 | 3.57 |
|  | 10 | OnC | 3.5 | 3.5 | 3.642 | 6 | 9.642 |
|  |  | OnI | 1.11 | 1 | 1.17 | 1.25 | 2.42 |
|  | 15 | OnC | 3.5 | 3.5 | 3.53 | 6 | 9.53 |
|  |  | OnI | 1.11 | 1 | 1.17 | 1.25 | 2.42 |
|  | 20 | OnC | 3.25 | 3.25 | 3.35 | 5.75 | 9.10 |
|  |  | OnI | 1.00 | 1 | 0.99 | 1.25 | 2.24 |

The results imply that the new method could get the better performance through learning the kernel matrix in the different parameters . The experimental results obtained on the facial database show that the verification performance of the new method is superior to that of the primary method Client Specific Kernel Discriminant Analysis (CSKDA). The classification based on imposter model is better than the client model. Moreover, there is a special relationship between FAR and FRR. When one larger, the other smaller. The method CSKDA need to choose a proper kernel function through many experiments, while the new method could learn the kernel from data automatically which could save a lot of time and have the robust performance.

5. Conclusion

The kernel function is introduced to solve the nonlinear pattern recognition problem. The advantage of a kernel method often depends critically on a proper choice of the kernel function. A promising approach is to learn the kernel from data automatically. Over the past few years, some methods which have been proposed to learn the kernel have some limitations: learning the parameters of some prespecified kernel function and so on. In this paper, the nonlinear face verification via learning the kernel matrix is proposed. The method CSKDA need to choose a proper kernel function through many experiments, while the new method could learn the kernel from data automatically which could save a lot of time and have the robust performance.